\documentclass[runningheads]{llncs}

\usepackage{eccv}

\usepackage{eccvabbrv}

\usepackage{graphicx}
\usepackage{booktabs}
\usepackage{multirow}
\usepackage{threeparttable}
\usepackage{pifont}
\usepackage{colortbl}
\usepackage{algorithm}
\usepackage{algpseudocode}
\usepackage{subcaption}
\usepackage{bbm}
\usepackage[accsupp]{axessibility}  % Improves PDF readability for those with disabilities.

\usepackage{hyperref}
\usepackage{microtype}

\usepackage{orcidlink}

\begin{document}

% no need to investigate
%\title{Mitigating the Model Forgetting Degradation in Long-Incremental 3D Indoor Object Detection}
% \title{Breaking the Forgetting Cycle in Long-Horizon Incremental 3D Object Detection}
\title{Breaking the Model Forgetting Cycle in Long-Incremental 3D Object Detection}

\titlerunning{Breaking the Model Forgetting Cycle in L-I3DOD}

\author{Peisheng Qian\inst{1, 2} \and
Jie Xu\inst{1} \and
Xulei Yang\inst{2,}$^{*}$ \and
Na Zhao\inst{1,}$^{*}$}
\authorrunning{P.~Qian et al.}
\institute{Singapore University of Technology and Design,\\
\email{\{jie\_xu2, na\_zhao\}@sutd.edu.sg} \and
Institute for Infocomm Research, A*STAR, Singapore, \\
\email{\{qian\_peisheng, yang\_xulei\}@a-star.edu.sg}}

% % TODO FINAL: Replace with your author list. 
% % Include the authors' OCRID for the camera-ready version, if at all possible.
% Second Author\inst{2,3}\orcidlink{1111-2222-3333-4444} \and
% Third Author\inst{3}\orcidlink{2222--3333-4444-5555}}

% % TODO FINAL: Replace with an abbreviated list of authors.

% % TODO FINAL: Replace with your institution list.
% Springer Heidelberg, Tiergartenstr.~17, 69121 Heidelberg, Germany

\maketitle
\renewcommand{\thefootnote}{}
\footnotetext{* Corresponding authors.}
\renewcommand{\thefootnote}{\arabic{footnote}}
\setcounter{footnote}{0}

\begin{abstract}
Incremental 3D object detection requires a detector to learn novel object classes while remembering previously learned ones over sequentially arriving data.
Previous methods, primarily based on pseudo-labeling, perform reasonably in short-incremental stages but still suffer from severe model forgetting when dealing with long-incremental sequences.
We investigate this failure and reveal a detrimental self-reinforcing cycle:
data distribution shift of novel classes causes model forgetting on old classes, which further produces accumulated error in pseudo-labeling that exacerbates model degradation.
To address this issue, we draw inspiration from the human learning process and propose the \emph{Learning-Dynamics-driven Memory and Review} (LDMR) framework.
LDMR monitors per-class detection quality at periodic training checkpoints and uses these learning-dynamics signals to drive two innovative mechanisms, namely 
(i) \emph{human-like intra-stage review} that divides each incremental stage into multiple sub-stages' training and concentrates on remembering the most-forgotten objects, and
(ii) \emph{scene-aware cross-stage memory evolution} that evolves a memory bank to transfer knowledge between two consecutive stages by jointly considering scene learnability and diversity.
Extensive experiments across multiple long-incremental protocols on indoor benchmarks SUN RGB-D and ScanNetV2 show that LDMR substantially mitigates the model forgetting and outperforms all baselines by a clear margin.
Code is available at \url{https://github.com/qianpeisheng/LDMR}.
\end{abstract}

\begin{figure}[!ht]
    \centering
    \begin{subfigure}[t]{0.69\linewidth}
        \centering
        \includegraphics[width=\linewidth]{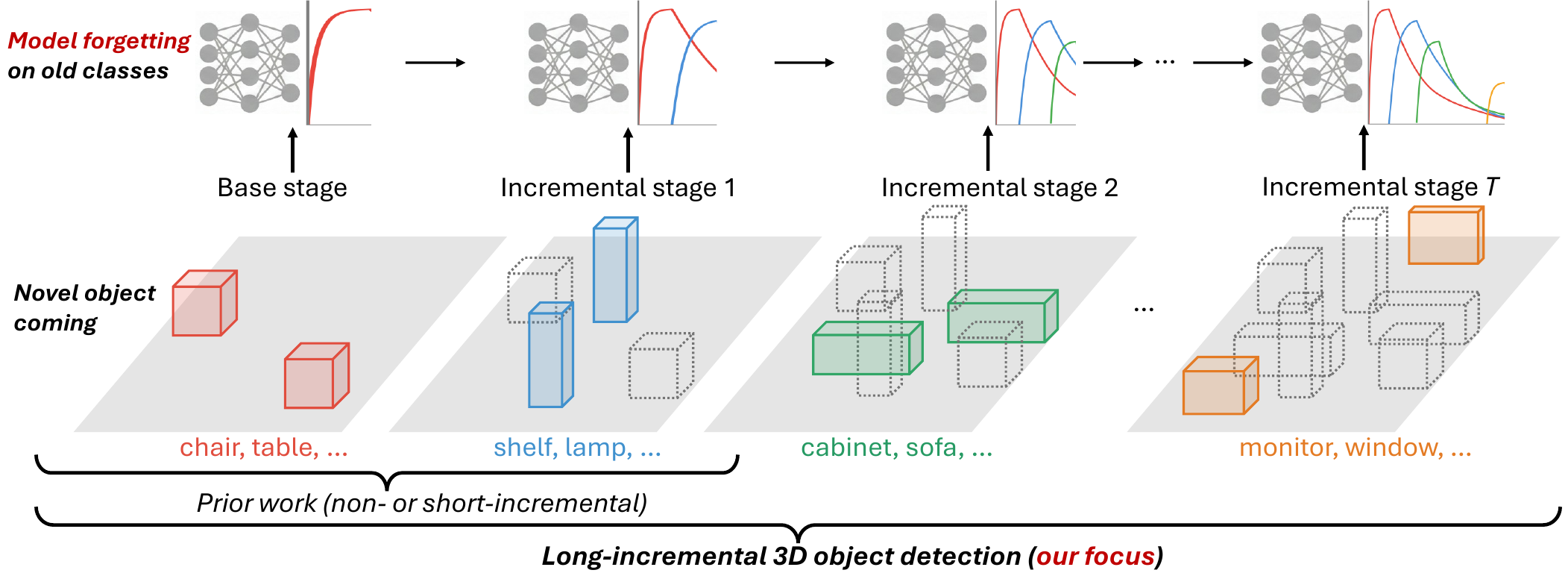}
        \caption{} % shows only (a)
        \label{fig:fig1a}
    \end{subfigure}
    \hfill
    \hfill
    \begin{subfigure}[t]{0.3\linewidth}
        \centering
        \includegraphics[width=\linewidth]{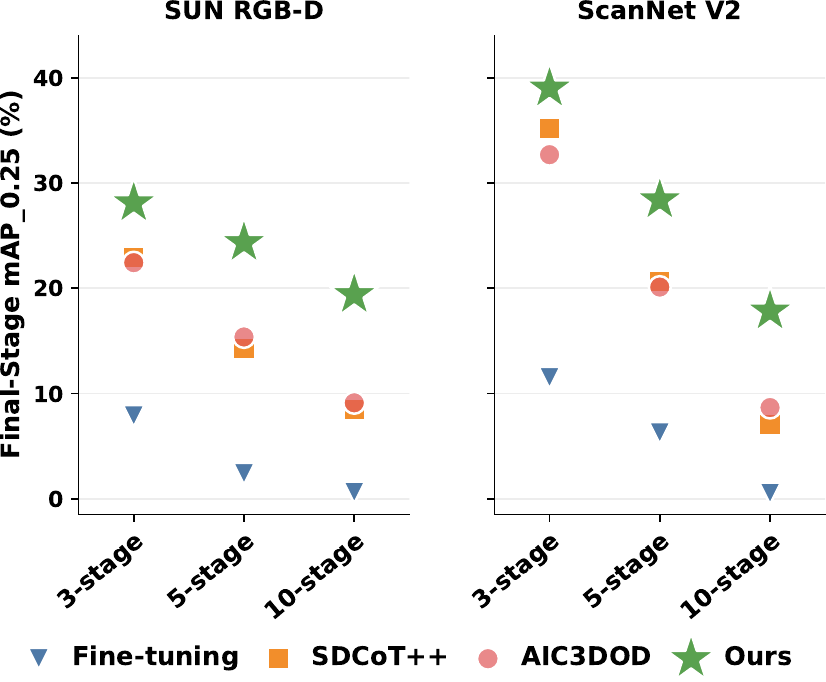}
        \caption{} % shows only (b)
        \label{fig:fig1b}
    \end{subfigure}
% \vspace{-0.3cm}
\caption{Task illustration and result comparison. (a) Long-incremental 3D object detection has the continuous incremental learning stages, where novel class annotations are available at the current stage; objects from previously learned old classes are unlabeled in the later stages. (b) Performance comparison over multiple long-incremental 3D detection tasks, showing our method consistently outperforms baseline approaches.}
\label{fig:fig1}
% \vspace{-0.4cm}
\end{figure}

\section{Introduction}
\label{sec:intro}

Indoor 3D object detection~\cite{zhao2022sdcot, zhao2024sdcot, aic3dod, jiao2024unlocking, zhao2020sess, al_indoor3d, zhu2026few, han2024dual} is a core capability for many applications such as augmented reality, embodied agents, and assistive robotics, where a system must localize and recognize objects from RGB-D scans or point clouds.
Recently, modern 3D detectors have advanced rapidly due to deep learning~\cite{qi2019votenet,tr3d2023, shen2024vdetr}, and their training processes typically assume a closed-set label space and a stationary training distribution.
In realistic deployments, however, new object classes and new environments usually arrive over time, and repeatedly retraining from scratch is often impractical~\cite{zhao2022sdcot,zhao2024sdcot, wu2026ccf, yuan2026graph, Xu_2026_CVPR}.

To address this limitation, \emph{Incremental 3D Object Detection} (I3DOD) has been proposed and has attracted increasing attention~\cite{zhao2022sdcot, zhao2024sdcot}.
As shown in Fig.~\ref{fig:fig1a}, I3DOD updates the detector over a sequence of incremental stages, each introducing novel classes while requiring the model to remember competence on all previously learned classes.
To achieve this, previous work mainly leverages pseudo-labeling strategies~\cite{zhao2022sdcot,zhao2024sdcot,aic3dod} to incrementally train detector models.
For example, SDCoT~\cite{zhao2022sdcot} introduces a teacher-student I3DOD framework in which a static teacher (the previous-stage model) generates pseudo labels for old classes, while a dynamic teacher (an exponential moving average of models) transfers underlying knowledge from the new data to the student network via consistency regularization.
Building on this, SDCoT++~\cite{zhao2024sdcot} consolidates pseudo labels from both teachers and calibrates class probabilities according to object occurrence frequencies, yielding more balanced and complete pseudo annotations that better handle the class imbalance inherent in 3D detection datasets.
More recently, AIC3DOD~\cite{aic3dod} advances the detection backbone with a point transformer architecture for stronger feature representation and higher-quality pseudo labels, and further incorporates room layout constraints across incremental stages.

Although previous I3DOD methods have achieved remarkable progress, they were typically designed for short-incremental stages, \textit{e.g.}, one or two incremental stages~\cite{zhao2022sdcot,zhao2024sdcot,aic3dod}.
When handling the 3D detection tasks with \emph{long-incremental} stages, these methods still fail to balance learning novel classes and remembering old knowledge, leading to severe model forgetting and performance degradation on old classes (see results in Fig.~\ref{fig:fig1b} and analysis in Sec.~\ref{sec:pilot_investigate}).
Handling many incremental stages is the more realistic and practically relevant setting: embodied agents or robotic assistants deployed over long periods will continuously encounter new object categories rather than just one or two updates. Motivated by this real-world requirement, we introduce \emph{Long-Incremental 3D Object Detection} (L-I3DOD), a formulation that studies 3D detection under prolonged streams of category increments and aims to enable lifelong learning for practical 3D perception systems.

In this paper, we investigate the model forgetting by theoretical and empirical analysis, and reveal a self-reinforcing cycle that raises the crucial challenge for L-I3DOD.
Specifically, class distribution shifts between novel and old stages produce optimization gradient misalignment,
which increases the model's empirical risk on old‑stage classes and leads to the model degradation on old-classes.
Moreover, the pseudo-labeling strategies conducted on old-classes inevitably introduce error that will accumulate and hinder the model's training over all incremental stages.
Motivated by {the human's learning process of ``learn--evaluate--review'',
\ie, the learned knowledge is evaluated by tests to identify which parts have been forgotten and then these forgotten parts are reviewed again,
we propose an innovative \emph{Learning-Dynamics-driven Memory and Review} (LDMR) framework to mitigate the model forgetting in long-incremental 3D object detection.
Concretely, we define the learning dynamics that monitors per-class detection quality at periodic training checkpoints,
and LDMR uses these learning-dynamics signals to drive two iterative processes: \emph{intra-stage review} and \emph{cross-stage memorization}. The intra-stage review process divides each incremental stage into multiple sub-stages' training and leverages the human-like learn--evaluate--review mechanism to remember the easily forgotten objects.
The cross-stage memorization process focuses on transferring knowledge between two consecutive stages by evolving a memory bank that jointly considers scene-aware learnability and diversity.

In summary, the contributions of this work are as follows:
\begin{itemize}
  \item This study aims to address the model forgetting challenge for L-I3DOD, where we deeply investigate the crucial issues and propose an innovative L-I3DOD methodology motivated by the natural human learning mechanism.
  \item We present a learning-dynamics-driven framework that unifies the human-like intra-stage review and scene-aware cross-stage memory evolution, under meaningful measurable retention signals with a limited memory budget.
  \item Extensive experiments, on SUN RGB-D and ScanNetV2 across different settings of 3/5/10-incremental stages, demonstrate our method can mitigate the model forgetting and achieve consistent improvements over prior methods.
\end{itemize}

\section{Related Work}
\label{sec:related}

\subsection{3D Object Detection}
\label{sec:related_3dod}
3D object detection predicts oriented 3D bounding boxes from point clouds reconstructed from RGB-D scans or captured by laser scanners~\cite{rukhovich2022fcaf3d, kolodiazhnyi2025unidet3d, lazarow2025cubify, cao2026vggt, lemeshko2026zoo3d, sheng2025ct3d++}.
Representative indoor pipelines include vote-based proposal generation~\cite{qi2019votenet},
transformer-style set prediction with object queries~\cite{det3d_3detr},
sparse-voxel two-stage detectors that refine proposals with geometric aggregation~\cite{wang2022cagroup3d},
and proposal-free transformers that directly aggregate global context~\cite{groupfree}.
Recent work further focuses on improving 3D model efficiency in feature aggregation for cluttered indoor scenes~\cite{tr3d2023} or exploring 2D foundation model based 3D detection~\cite{jiao2024unlocking}.
In this work, we adopt TR3D~\cite{tr3d2023} and VoteNet~\cite{qi2019votenet} as backbone detectors.

Since 3D box annotation for indoor scenes is expensive, a growing line of work studies label-efficient indoor detection, including weakly-supervised learning~\cite{prompt3d}, semi-supervised teacher--student training~\cite{zhao2020sess}, and sparse-supervised learning via class-level prototypes~\cite{Zhu_2025_CVPR_CPDet3D}. Active learning has also been introduced to select informative scenes for annotation under a fixed labeling budget~\cite{al_indoor3d}. Our cross-stage memory evolution (Sec.~\ref{sec:memory}) shares the spirit of selecting high-value scenes under a budget, but targets the incremental learning setting.

\subsection{Class-Incremental Learning}
\label{sec:related_cil}
Class-incremental learning (CIL) studies continual updates where new classes arrive over time while performance on previously learned classes must be preserved.
Common approaches include parameter-importance regularization (e.g., EWC-style constraints)~\cite{cil_ewc},
distillation-based retention (e.g., LwF-style feature/logit matching)~\cite{cil_lwf},
and rehearsal via exemplar memory~\cite{cil_icarl}.
More recent directions emphasize parameter-efficient adaptation such as prompt- and adapter-based continual learning~\cite{cil_prompt, coda_prmopt, wu2025sdlora, NEURIPS2025_1663fba7, savadikar2026cheem}; however, these methods are primarily designed for image classification and do not address the localization objectives and missing-annotation issues that arise in detection.

Class-incremental object detection is particularly challenging due to dense foreground-background imbalance, structured localization objectives, and missing annotations for previously learned classes, where unlabeled old objects can be mistakenly optimized as background~\cite{liu2023abr, liu2023cldetr}.
Prior incremental {2D} detection methods commonly rely on knowledge distillation and replay strategies to address this~\cite{incdet_2d_ilod, peng2020faster_ilod, liu2023abr, feng2022erd, liu2023cldetr}.
These 2D strategies, however, do not transfer directly to 3D indoor detection, where old-class and novel-class objects physically co-exist within the same scenes, intensifying the supervision conflict.
Notably, despite rehearsal being well-established in CIL~\cite{cil_icarl, cil_survey}, no prior work has explored memory-bank-based replay for incremental 3D object detection.

\subsection{Incremental 3D Object Detection}
\label{sec:related_i3dod}
Incremental 3D object detection (I3DOD) incrementally introduces novel categories while old-stage data is unavailable; in practice, old objects may still appear but remain unlabeled, creating stage-wise supervision conflicts~\cite{yun2021conflicts}.
Most existing methods adopt teacher-student training with pseudo labels and distillation.
For example, SDCoT proposes static-dynamic co-teaching, where a frozen teacher transfers old-class knowledge and a dynamic teacher stabilizes learning on new-stage data~\cite{zhao2022sdcot}.
SDCoT++ improves co-teaching by consolidating teacher pseudo labels and calibrating class scores to mitigate imbalance and missing old objects~\cite{zhao2024sdcot}.
CoCo-Teach provides another co-teaching variant~\cite{coco}.
AIC3DOD incorporates transformer-based representations and room-layout constraints to regularize incremental training under incomplete annotations~\cite{aic3dod}.
Across all these methods, pseudo-labeling is the most effective component for retaining old-class knowledge. In long-incremental settings, however, even pseudo labels become unreliable as errors compound over stages (Sec.~\ref{sec:pilot_investigate}).

Existing I3DOD methods~\cite{zhao2022sdcot,zhao2024sdcot,aic3dod, in_defense} mitigate forgetting through pseudo-labels or regularization techniques, without replaying real annotated old-class scenes.
We draw on the fact that humans' learning relies on note-taking and use memory bank mechanism in long-incremental detection tasks.
Note that memory bank is a common practice for incremental learning~\cite{cil_icarl, liu2023abr}, but as of yet, no specific methods have been proposed for I3DOD domain.
What is more crucial is how to define the memory bank and the update rules for I3DOD because some information in 3D scenes might be misleading or redundant for incremental learning.
In this work, we jointly consider the scene diversity and learnability to establish an innovative scene-aware cross-stage memory evolution method. Moreover, we propose the human-like intra-stage review and show that they mitigate model forgetting over long-incremental 3D detection tasks.

\section{Background and Analysis}

\subsection{Problem Definition}
\label{sec:prob_def}
\textbf{Notation.}
Let $\mathcal{C}$ be the full class set, partitioned into $T$ disjoint groups
$\{\mathcal{C}_t\}_{t=1}^{T}$, \textit{i.e.}, $\mathcal{C}_i \cap \mathcal{C}_j=\emptyset$ for $i\neq j$ and
$\bigcup_{t=1}^{T}\mathcal{C}_t=\mathcal{C}$.
We use $\mathcal{C}_{\le t}$ and $\mathcal{C}_{<t}$ to denote the
seen classes up to and before stage $t$, respectively.
At each stage $t$, we call $\mathcal{C}_t$ the \emph{novel} classes and $\mathcal{C}_{<t}$ the \emph{old} classes.
The learner receives a stage-specific dataset $\mathcal{D}_t=\{\mathcal{X}_t,\mathcal{Y}_t\}$, where $x \in \mathcal{X}_t$ is an indoor 3D scene
and $y_t \in \mathcal{Y}_t$ contains 3D object detection annotations only from the novel classes $\mathcal{C}_t$.

\noindent\textbf{Incremental 3D object detection (I3DOD).}
Training of I3DOD proceeds over stages $t=1,\dots,T$. It trains $f_{\theta_1}$ on $\mathcal{D}_1$ and then for $t\ge 2$ updates $f_{\theta_{t-1}}$ to $f_{\theta_t}$ using only $\mathcal{D}_t$, where old-class instances may appear but are unlabeled. After stage $t$, the detector must detect all seen classes $\mathcal{C}_{\le t}$.
Prior work~\cite{zhao2022sdcot,zhao2024sdcot, aic3dod} typically consider short-incremental stages' I3DOD tasks (\textit{e.g.}, $T{=}2$ or $T{=}3$).
In this study, we focus on the long-incremental stages' I3DOD tasks (\textit{e.g.}, $T \geq 3$).

\subsection{Theoretical and Empirical Analysis of Model Forgetting}
\label{sec:pilot_investigate}

\begin{figure}[!t]
  \centering
  \begin{subfigure}[t]{0.32\linewidth}
    \centering
    \includegraphics[width=\linewidth]{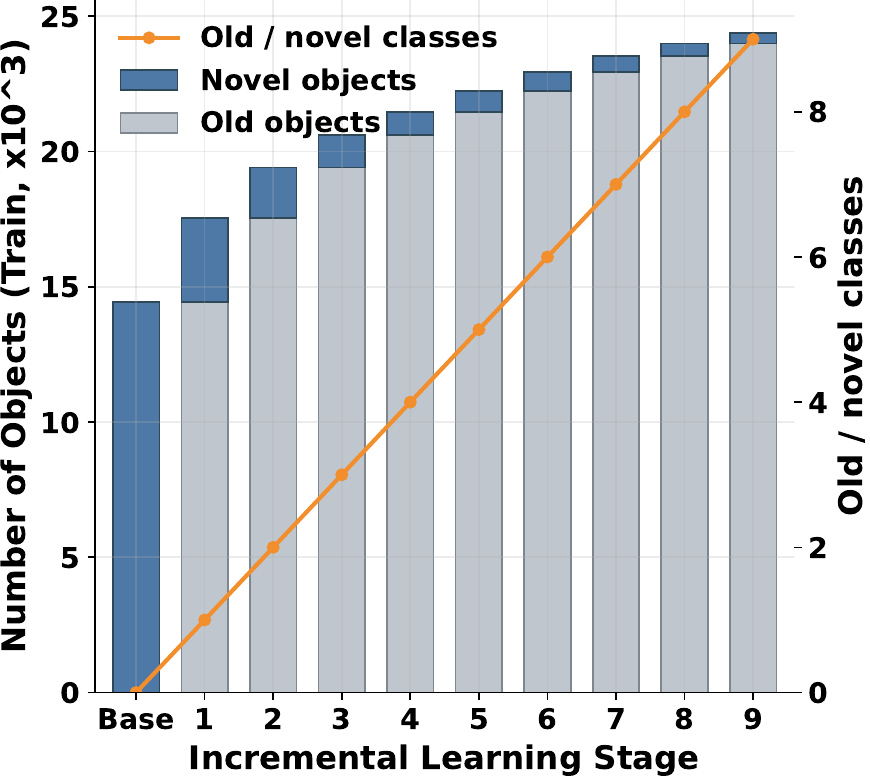}
    \caption{Training data variation.}
    \label{fig:pilot_figure_a}
  \end{subfigure}\hfill
  \begin{subfigure}[t]{0.32\linewidth}
    \centering
    \includegraphics[width=\linewidth]{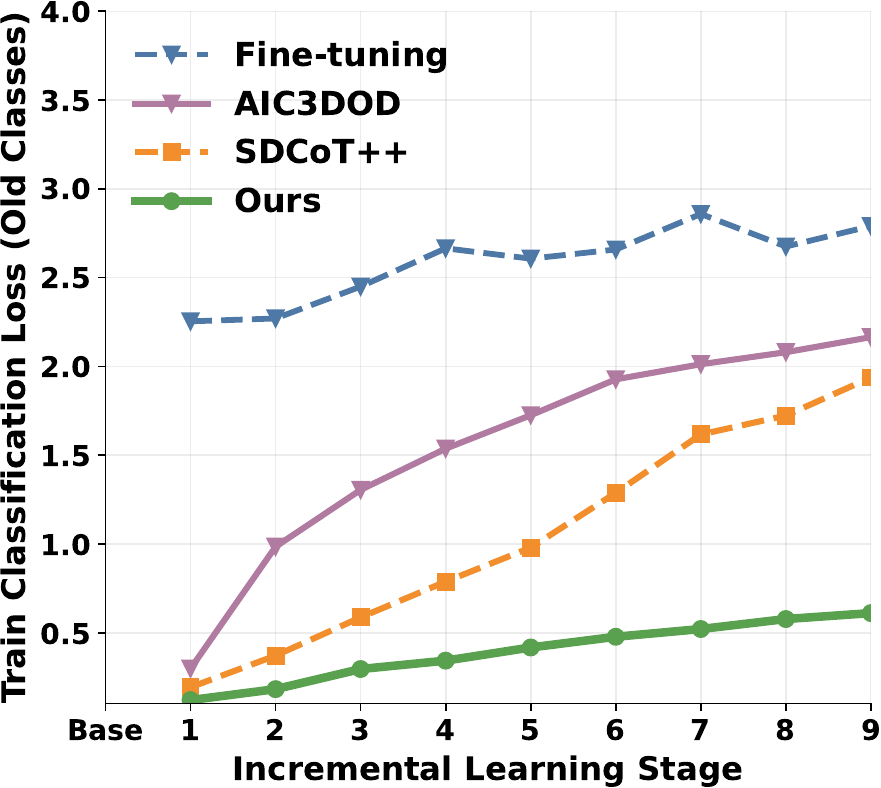}
    \caption{Evaluated training loss.}
    \label{fig:pilot_figure_b}
  \end{subfigure}\hfill
  \begin{subfigure}[t]{0.32\linewidth}
    \centering
    \includegraphics[width=\linewidth]{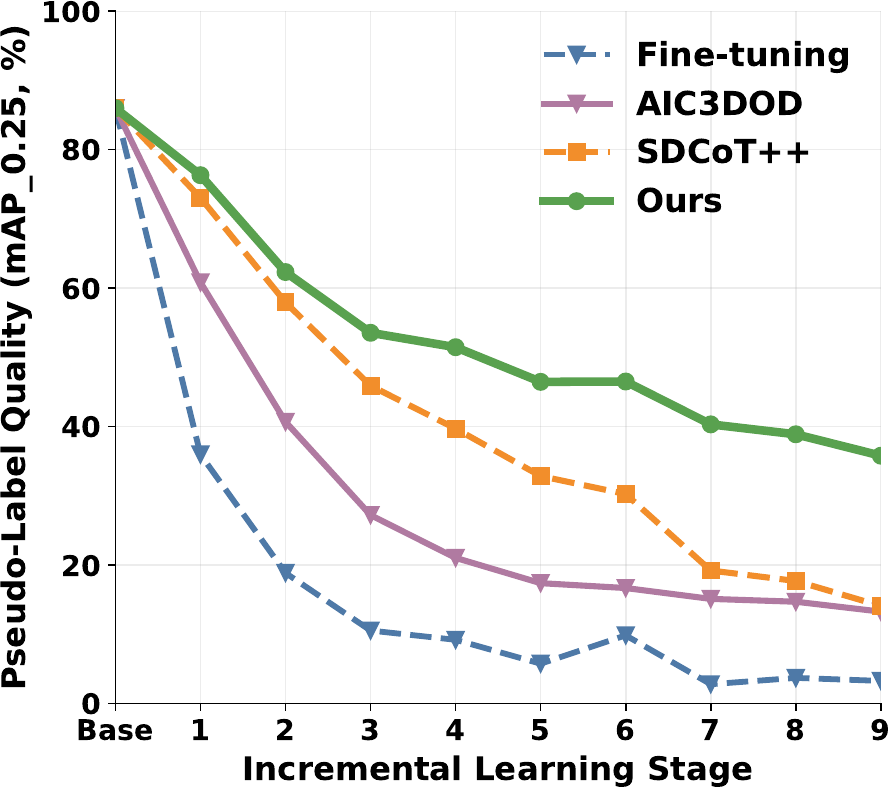}
    \caption{Pseudo-label quality.}
    \label{fig:pilot_figure_c}
  \end{subfigure}
% \vspace{-0.1cm}
\caption{
Empirical analysis of L-I3DOD on SUN RGB-D (40 classes).
(a) Training data variation indicates the distribution shift over long-incremental stages.
(b) Training loss or empirical risk evaluated on old classes at each stage.
(c) The quality of pseudo-labels generated by each stage's model.
% The results reveal distribution imbalance and progressively degraded pseudo labels as key causes of forgetting.
The results reveal that the vanilla Fine-tuning strategy and the I3DOD methods SDCoT++~\cite{zhao2024sdcot} and AIC3DOD~\cite{aic3dod} still face the heavy model forgetting in L-I3DOD tasks, while our method achieves significant advancement.}
\label{fig:pilot_figure}
% \vspace{-0.1cm}
\end{figure}

Firstly, we conduct a pilot study to diagnose the model forgetting in long-incremental 3D object detection (L-I3DOD).
As shown in Fig.~\ref{fig:pilot_figure}, we observe that (i) distribution shift and (ii) error accumulation tend to be the crucial factors causing the heavy model forgetting for L-I3DOD tasks.

\noindent\textbf{I) Distribution shift issue:} training on the shifted data distribution of novel classes leads to the model degradation on old classes.
Firstly, we give the following theoretical analysis to illustrate this. Let stages $t=1,\dots,T$ have data distributions $P_t$ on sample space $\mathcal{Z}=\mathcal{X}\times\mathcal{Y}$.  
Let the model parameter be $\theta\in\mathbb{R}^d$ and define the stage-$t$ expected detection loss as $\mathcal{L}_t(\theta)=\mathbb{E}_{z\sim P_t}\big[\ell(f_\theta,z)\big]$,
and the expected gradient is $g_t(\theta)=\nabla_\theta \mathcal{L}_t(\theta)=\mathbb{E}_{z\sim P_t}[\nabla_\theta \ell(f_\theta,z)]$.
We leverage the Wasserstein distribution distance to establish the long-incremental stages' forgetting upper bound as follows (the proof is given in the supplementary material):
\begin{theorem}[Long-incremental stages' forgetting upper bound]
Let $i\le s$ be an index of a previously trained stage.
Perform gradient-descent updates at stages $u=s+1,\dots,t$ with step sizes $\{\eta_u\}$ and we have $\theta_u=\theta_{u-1}-\eta_u\, g_u(\theta_{u-1})$, where the stage-$u$ population gradient is
$g_u(\theta_{u-1})=\mathbb{E}_{z\sim P_u}\big[\nabla_\theta \ell(f_\theta,z)\big]\Big|_{\theta=\theta_{u-1}}$.
Under the regularity conditions stated in the proof, the cumulative forgetting (increase of expected loss for stage $i$) after updates up to stage $t$ satisfies
\begin{equation}
\mathcal{L}_i(\theta_t)-\mathcal{L}_i(\theta_s)
\le \sum_{u=s+1}^t\Big( 
\eta_u K\,W_1(P_i,P_u)\|g_u(\theta_{u-1})\| 
+ \tfrac{L\eta_u^2}{2}\|g_u(\theta_{u-1})\|^2
\Big)
\label{eq:loss_bound}
\end{equation}
where $W_1(P_i,P_u)$ is the 1‑Wasserstein distance on $\mathcal{Z}$, $K$ is the directional Lipschitz constant and $L$ is the smoothness constant in parameter space, and each $\|g_u(\theta_{u-1})\|$ is the Euclidean norm of the population gradient evaluated at $\theta_{u-1}$.
\end{theorem}
This bound explains why long-incremental stage 3DOD forgets: distribution shifts between two stages (measured by \(W_1(P_i,P_u)\)) and large new‑stage gradients \(\|g_u(\theta_{u-1})\|\) produce gradient misalignment that increases old‑stage loss, while the input-gradient sensitivity (measured by \(K\)), optimization curvature smoothness (measured by \(L\)), and step size (\(\eta_u\)) further amplify this effect.

Secondly, the empirical evidence in Fig.~\ref{fig:pilot_figure} aligns our theoretical analysis, which suggests that
the difference in quantity between novel-class objects and old-class objects becomes large during the long-incremental training process (Fig.~\ref{fig:pilot_figure_a});
supervised learning on the minority of novel-class data results in increasing the model empirical risk (training loss) on the majority of old-class data distribution (Fig.~\ref{fig:pilot_figure_b}),
which indicates the occurrence of model forgetting in L-I3DOD.

\noindent\textbf{II) Error accumulation issue:} the error from pseudo-labeling strategies will accumulate and hinder the model over long-incremental stages.
To enhance remembering old classes, pseudo-labeling strategies~\cite{zhao2022sdcot,zhao2024sdcot,aic3dod} are widely adopted for re-labeling the old classes during incremental 3D object detection.
However, the signals generated by pseudo-labeling may inherently contain errors, and as the number of incremental stages increases, the accumulated errors become significant and the pseudo-label quality for the old classes will further deteriorate.
Taking the representative pseudo-labeling based I3DOD method SDCoT++~\cite{zhao2024sdcot} as an example, one can observe that the trained model has the gradually increasing empirical risk on old classes (Fig.~\ref{fig:pilot_figure_b}) and the decreasing pseudo-label quality (Fig.~\ref{fig:pilot_figure_c}).
As a result, these two issues lead to a self-reinforcing cycle: distribution shift degrades model performance on old classes and wrong pseudo labels accumulate errors over long-incremental stages, making existing methods insufficient to prevent model forgetting.
This motivates us to study how to effectively address the model forgetting challenge so as to make I3DOD applicable in practical long-incremental scenarios. 

\section{Method}
\label{sec:method}

% \subsection{Methodology Motivation and Overview}
\subsection{Overview}
\label{sec:pilot_motivation}
\begin{figure}[!t]
\centering
\includegraphics[width=\linewidth]{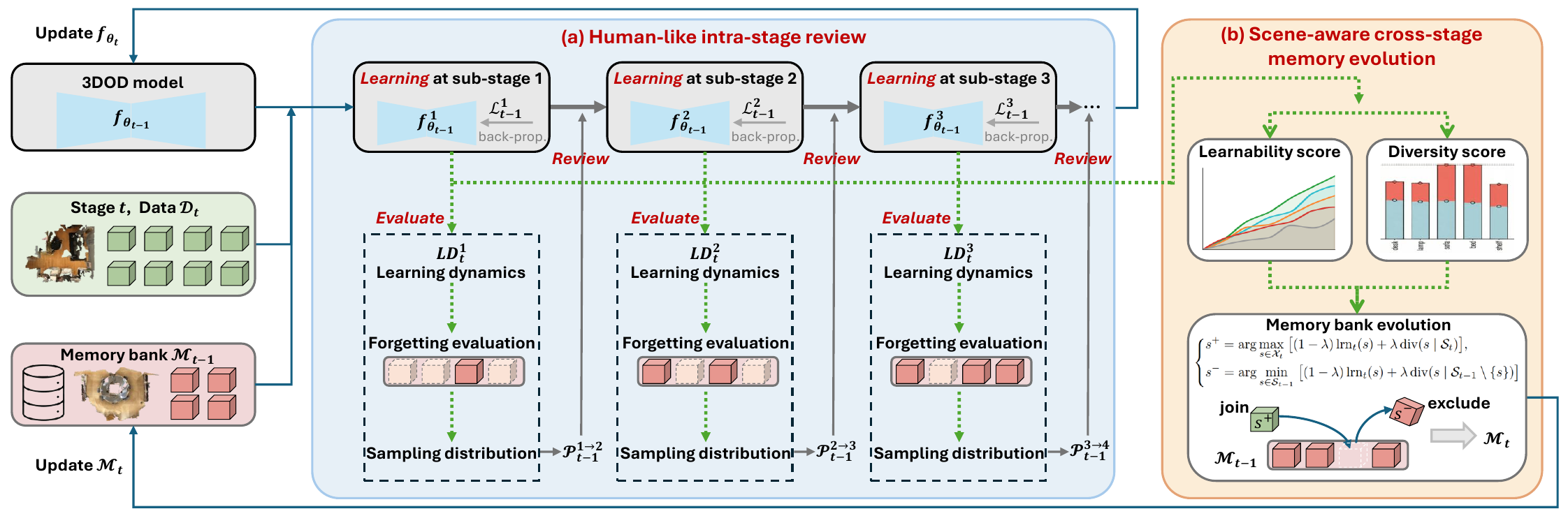}
% \vspace{-0.5cm}
\caption{\textbf{Method overview.} Given the $t$-th incremental stage data $\mathcal{D}_t$, our LDMR leverages the model ${f}_{\theta_{t-1}}$ and the memory bank $\mathcal{M}_{t-1}$ inherited from the last $(t-1)$-th stage, to conduct (a) intra-stage human-like review which simulates the learn--evaluate-review process by which humans overcome forgetting, to obtain the new model ${f}_{\theta_{t}}$.
Then, LDMR performs (b) scene-aware cross-stage memory evolution which establishes the learnability and diversity scores for achieving the cross-stage knowledge transfer, by updating the memory bank $\mathcal{M}_{t}$.
}
\label{fig:ldrr_overview}
% \vspace{-0.3cm}
\end{figure}
Motivated by the ``learn-evaluate-review'' process by which humans overcome forgetting, we propose the {Learning-Dynamics-driven Memory and Review} (LDMR) framework to address the model forgetting in long-incremental 3D detection. The framework is shown in Fig.~\ref{fig:ldrr_overview}. It consists of the~\emph{human-like intra-stage review} and~\emph{scene-aware cross-stage memory evolution}.
% \textbf{I) Scene-aware cross-stage memory evolution.}
The technical details are presented in the following sections.

\subsection{Human-like Intra-stage Review}
\label{sec:review}

As Sec.~\ref{sec:pilot_investigate} reveals, training on the $t$-th stage data $\mathcal{D}_t$ focuses on fitting novel classes $\mathcal{C}_{t}$ and will introduce model forgetting on old classes $\mathcal{C}_{<t}$ due to the distribution shift between $\mathcal{D}_{t}$ and $\mathcal{D}_{<t}$.
To address this issue, we design the human-like intra-stage review that maximally learns novel classes while remembering old ones, given the model parameter $f_{\theta_{t-1}}$ and memory bank $\mathcal{M}_{t-1}$ of the last stage.
First, we formally give the definition of memory bank $\mathcal{M}_{t-1}$ as follows:

\noindent\textbf{Memory bank.}
The memory bank aims to support the knowledge transfer from the previous $t-1$ stages to the current $t$-th stage.
Hence, the memory bank $\mathcal{M}_{t-1}$ preserves a small set of annotated scenes from historical data set $\mathcal{D}_{< t}$, by
\begin{equation}
\mathcal{M}_{t-1} = \{\mathcal{S}_{t-1},\mathcal{Y}|_{\mathcal{S}_{t-1}}\},~~\mathcal{S}_{t-1} \subset \mathcal{X}_{< t},~~\mathcal{Y}|_{\mathcal{S}_{t-1}} \subset \mathcal{Y}_{< t},~~|\mathcal{S}_{t-1}| = B,
\end{equation}
where $B$ is a fixed-budget size, $\mathcal{X}_{< t}$ denotes the historical scenes and $\mathcal{Y}_{< t}$ includes the annotations from the previous stages.
For scenes $\mathcal{S}_{t-1}$ in the memory bank, $\mathcal{Y}|_{\mathcal{S}_{t-1}}$ is the corresponding annotation set whose classes belong to $\mathcal{C}_{< t}$.
$\mathcal{M}_{0} = \emptyset$.
Then, to simulate the human's learning process, we divide one incremental stage's model training $f_{\theta_{t-1}} \rightarrow f_{\theta_{t}}$ into $I$ sub-stages as $f^1_{\theta_{t-1}}, f^2_{\theta_{t-1}}, f^3_{\theta_{t-1}},..., f^I_{\theta_{t-1}}$ (finally we have $f_{\theta_{t}} = f^I_{\theta_{t-1}}$), {where $f^i_{\theta_{t-1}}$ is a sub-stage model checkpoint}.
Before and after the sub-stage's training from $f^i_{\theta_{t-1}}$ to $f^{i+1}_{\theta_{t-1}}$, we record the \emph{learning dynamics} which is defined as follows:

\noindent\textbf{Learning dynamics.}
To dynamically monitor the model's detection performance on old classes, the learning dynamics is defined as a set of the per-class recall performance on the memory bank $\mathcal{M}_{t-1}$: 
\begin{equation}
LD^i_t = \{R^i_t(s,c)\},~~{s\in \mathcal{S}_{t-1}},~~{c\in \mathcal{Y}|_{\mathcal{S}_{t-1}}},
\label{eq:ld_t}
\end{equation}
where $R^i_t(s,c)$ is the recall of the class $c$ in the scene $s$ in the sub-stage $i$.

\noindent\textbf{Forgetting evaluation.}
Furthermore, we evaluate the two learning dynamics $LD^i_t$ and $LD^{i+1}_t$ to check the forgetting knowledge on old classes from model $f^i_{\theta_{t-1}}$ to $f^{i+1}_{\theta_{t-1}}$.
The per-class performance drop for each scene $s$ and old class $c \in \mathcal{C}_{<t}$ can be computed by $\max(0,\, LD^i_t({s,c}) - LD^{i+1}_t({s,c}))$.
In this way, we obtain the following scene-level forgetting metric:
\begin{equation}
F_s^{i\to i+1} =
\sum_{c\in\mathcal{C}_{<t}} \log(1 + n_{s,c}) \times \max\left(0,\, R^i_t({s,c}) - R^{i+1}_t({s,c})\right),
\label{eq:scene_drop}
\end{equation}
where $\log(1 + n_{s,c})$ is a log-scaled object-count weight and $n_{s,c}$ denotes the number of ground-truth instances of class $c$ in scene $s$.

Scenes with larger $F_s^{i\to i+1}$ indicate that the model has a more severe forgetting of the objects in those scenes during the sub-stage's training.
This guides us to carry out the following intra-stage review process.

\noindent\textbf{Intra-stage review with sampling distribution.}
We leverage the scene-level forgetting metrics $\{F_s^{i\to i+1}\}$, {$s\in \mathcal{M}_{t-1}$} to guide
the model to emphasize the forgotten knowledge in the next sub-stage's training (\textit{i.e.}, from $f^{i+1}_{\theta_{t-1}}$ to $f^{i+2}_{\theta_{t-1}}$).

To achieve this, we construct a unified sampling pool from the stage data and the memory bank $\mathcal{X}_t \cup \mathcal{S}_{t-1}$, where each scene is assigned an adaptive sampling weight.
To be specific, the sampling weight of scenes in the current stage data $\mathcal{X}_t$ is fixed at $w_s^{i\to i+1} = 1$, and that in the memory bank $\mathcal{S}_{t-1}$ incorporates a forgetting-proportional boost by the following equation:
\begin{equation}
w_s^{i\to i+1} = 1 + \eta \cdot F_s^{i\to i+1} \cdot \mathbb{I}(s \in \mathcal{S}_{t-1}),
\label{eq:replay_weight}
\end{equation}
where $\eta \ge 0$ controls the strength of the review emphasis. $\mathbb{I}(\cdot)$ is an indicator function.
Then, the probabilistic sampling distribution for the intra-stage review process of sub-stage $i$ to sub-stage $i+1$ over the unified pool is given by
\begin{equation}
\mathcal{P}_{t-1}^{i\to i+1}(s) = \frac{w_s^{i\to i+1}}{\sum_{j \in \mathcal{D}_t \cup \mathcal{M}_{t-1}} w_j^{i\to i+1}}.
\label{eq:replay_prob}
\end{equation}

At the next sub-stage from $f^{i+1}_{\theta_{t-1}}$ to $f^{i+2}_{\theta_{t-1}}$, training scenes are drawn from $\mathcal{D}_t \cup \mathcal{M}_{t-1}$ according to $\mathcal{P}_{t-1}^{i\to i+1}(s)$ and the detector is optimized with:
\begin{equation}
\mathcal{L}_{t-1}^{i+1} = \mathbb{E}_{s \sim \mathcal{P}_{t-1}^{i\to i+1}(s)}\left[\mathcal{L}_{\mathrm{det}}\!\left(s;\; \theta_{t-1}\right)\right],
\label{eq:loss}
\end{equation}
where $\mathcal{L}_{\mathrm{det}} = \mathcal{L}_{\mathrm{cls}} + \mathcal{L}_{\mathrm{bbox}}$ combines the classification and bounding-box regression losses following prior 3D detection work~\cite{tr3d2023, qi2019votenet}.

\subsection{Scene-aware Cross-stage Memory Evolution}
\label{sec:memory}
After stage-$t$ training completes, we compute the learning dynamics $\{LD^i_t\}_{i=1}^I$ for both old and novel classes on the data set $\mathcal{D}_{t} \cup \mathcal{M}_{t-1}$, generalizing Eq.~\ref{eq:ld_t} to all classes:
\begin{equation}
LD^i_t = \{R^i_t(s,c)\},~~{s\in \mathcal{X}_{t}}\cup\mathcal{S}_{t-1},~~{c\in \mathcal{C}_{t}}\cup\mathcal{Y}|_{\mathcal{S}_{t-1}},
\label{eqLD}
\end{equation}
which is leveraged to establish the scene-aware cross-stage memory evolution for updating the memory bank, \textit{i.e.}, achieving $\mathcal{M}_{t-1}\to \mathcal{M}_{t}$.

The key insight is that the memory bank should prioritize the learnable and representative scenes, \textit{i.e.}, {those where the model has shown it can improve with exposure but has not yet retained that improvement},
and the scene objects have the diversity to represent the entire data distribution.
Therefore, we define the following \textit{learnability} and \textit{diversity} scores in our memory bank evolution strategy.

\noindent\textbf{Learnability score.} To quantify how learnable the scene's objects are, we define the scene-aware learnability score for $s \in \mathcal{X}_{t}\cup\mathcal{S}_{t-1}$ as follows:
\begin{equation}
\mathrm{lrn}_t(s) = \sum_{c} \omega_t(c) \cdot \log(1 + n_{s,c}) \cdot G_t(s,c),
\label{eq:scene_learnability}
\end{equation} 
where $\omega_t(c)$ we introduced is an under-learning metric for class $c$, and $\log(1 + n_{s,c})$ is a log-scaled object-count weight.
$G_t(s,c)$ measures the learning gain of scene $s$ on class $c$ to reflect the learnable metric.

To obtain $\omega_t(c)$ for each class $c \in {\mathcal{C}_{t}}\cup\mathcal{Y}|_{\mathcal{S}_{t-1}}$, we base on $\{LD^i_t\}_{i=1}^I$ in Eq.~(\ref{eqLD}) to compute the class-level recall of the $i$-th sub-stage across scenes $R^i_{t}(c)=1/S \cdot \sum_{s} R_t^i(s,c)$, $S=|{\mathcal{X}_{t}}\cup\mathcal{S}_{t-1}|$, and record the peak class-level recall $\hat R^i_{t}(c) = \text{max}\{R^1_{t}(c),R^2_{t}(c),\dots,R^I_{t}(c)\}$ and class-level recall $R^I_{t}(c)$ for the final sub-stage.
A class is under-learnt when:
(i) it remains under-detected ($R^I_{t}(c)$ is low),
(ii) its quality has dropped from peak ($\hat R^i_{t}(c) > R^I_{t}(c)$), and
(iii) it is learnable ($\hat R^i_{t}(c)$ is not close to zero).
These derive our under-learning metric for class $c$:
\begin{equation}
\omega_t(c)
= \left[\left(1 -R^I_{t}(c)\right) + \max\left(0,\, \hat R^i_{t}(c) - R^I_{t}(c)\right)\right] \cdot \hat R^i_{t}(c).
\label{eq:class_need}
\end{equation}

To obtain $G_t(s,c)$, we summarize the positive recall gains of class $c$ in scene $s$ between consecutive sub-stages by:
\begin{equation}
G_t(s,c) = \sum\limits_{i=2}^{I} \max\left(0,\, R^i_t(s,c) - R^{i-1}_t(s,c)\right).
\label{eq:learning_gain}
\end{equation}
This metric reflects that objects of class $c$ in scene $s$ can achieve how much performance gain during the increment stage $t$.

In this way, our learnability score in Eq.~(\ref{eq:scene_learnability}) not only reflects class-level under-learning information via $\omega_t(c)$, but also captures scene-level possible learning gain via $G_t(s,c)$.

\noindent\textbf{Diversity score.}
Beyond learnability, we further consider scene diversity to ensure the memory bank covers a broad range of learning patterns. To this end, we embed each scene by its learning dynamics pattern and favor candidates that differ from scenes already in the memory bank.

Specifically, for each scene $s \in \mathcal{X}_t \cup \mathcal{S}_{t-1}$, we construct a two-dimensional embedding $\mathbf{e}_t(s) \in \mathbb{R}^2$ from its learning dynamics.
The first dimension aggregates the class-weighted learning gain $\sum_c \omega_t(c) \cdot G_t(s,c)$, and the second aggregates the class-weighted quality drop $\sum_c \omega_t(c) \cdot D_t(s,c)$, where $D_t(s,c) = \hat R^i_{t}(s, c) - R^I_t(s,c)$ is the drop from peak to final recall.
As a result, we have
\begin{equation}
\mathbf{e}_t(s) = \left[\sum_c \omega_t(c) \cdot G_t(s,c); \sum_c \omega_t(c) \cdot D_t(s,c)\right] \in \mathbb{R}^2.
\label{eq:embedding}
\end{equation}
Then, the embedding vector is L2-normalized to the unit circle and thus vectors of different scenes become comparable.
Scenes with similar $\mathbf{e}_t(s)$ exhibit similar learning dynamics patterns and they usually offer limited information gain.

Let $\mathcal{S}_{t}$ denote the set of scenes already selected for the $t$-th increment stage. The diversity score of a candidate $s$ against the memory bank is defined as:
\begin{equation}
\mathrm{div}(s \mid \mathcal{S}_{t})
= 1 - \frac{1}{|\mathcal{S}_{t}|}
  \sum_{j \in \mathcal{S}_{t}}
  \max\bigl(0,\;\mathbf{e}_t(s)^\top \mathbf{e}_t(j)\bigr),
\label{eq:diversity}
\end{equation}
where higher values indicate greater diversity from the existing memory bank.

\noindent\textbf{Memory bank evolution.}
We combine the scene-aware learnability score from Eq.~\eqref{eq:scene_learnability} and diversity score from Eq.~\eqref{eq:diversity} into a joint selection objective.
Our memory bank evolution consists of the following two update rules:
\begin{equation}
\left\{
\begin{aligned}
&s^+ = \arg\max_{s \in \mathcal{X}_{t}}\;
  \bigl[(1-\lambda)\,\mathrm{lrn}_t(s)
  + \lambda\,\mathrm{div}(s \mid \mathcal{S}_{t})\bigr], \\
&s^- = \arg\min_{s \in \mathcal{S}_{t-1}}\;
  \bigl[(1-\lambda)\,\mathrm{lrn}_t(s)
  + \lambda\,\mathrm{div}(s \mid \mathcal{S}_{t-1} \setminus \{s\})\bigr],
\end{aligned}
\right.
\label{eq:bank_evolution}
\end{equation}
where $\lambda \in [0,1]$ balances the learnability and diversity.
In Eq.~\ref{eq:bank_evolution}, the first rule identifies the most valuable candidate $s^+$ from the current stage data $\mathcal{X}_t$ to be added to the memory bank.
And the second rule identifies the least valuable scene $s^-$ in the existing memory bank $\mathcal{S}_{t-1}$ to be evicted.
The evicted scene $s^-$ is replaced by the candidate $s^+$.
This update process repeats until all $s \in \mathcal{X}_t$ have been checked whether it should be added to $\mathcal{S}_{t}$, and finally we achieve the memory bank evolution from $\mathcal{M}_{t-1} = \{\mathcal{S}_{t-1},\mathcal{Y}|_{\mathcal{S}_{t-1}}\}$ to $\mathcal{M}_{t} = \{\mathcal{S}_{t},\mathcal{Y}|_{\mathcal{S}_{t}}\}$.

\section{Experiments}
\label{sec:experiments}
\subsection{Experimental Setups}
\label{sec:exp_setup}

\noindent\textbf{Datasets.}
We adopt two indoor 3D object detection benchmarks to evaluate the proposed L-I3DOD task.
\textbf{SUN RGB-D}~\cite{Song_2015_CVPR} contains 10,335 RGB-D images captured by multiple sensors, with 5,285 training and 5,050 test samples.
\textbf{ScanNetV2}~\cite{dai2017scannet} contains 1,513 reconstructed RGB-D scans with 1,201 scans for training and 312 for evaluation.

\noindent\textbf{Incremental protocol.}
We evaluate under three incremental protocols with larger label spaces and more stages than prior work~\cite{zhao2022sdcot,zhao2024sdcot, aic3dod}.
For {SUN RGB-D}, we use 40 classes for 3-stage (20+10+10), 5-stage (5 stages of 8 classes each), and 10-stage (10 stages of 4 classes each).
For {ScanNetV2}, we use 35 classes for 3-stage (15+10+10), 5-stage (5 stages of 7 classes each), and 10-stage (10 stages of 3--4 classes each).
The classes are arranged in descending frequency order so that later stages introduce progressively rarer categories, making the task increasingly challenging.

\noindent\textbf{Implementation details.}
We evaluate with two 3D detector backbones, \ie, TR3D~\cite{tr3d2023} and VoteNet~\cite{qi2019votenet}.
We train the base stage for 90 epochs and each incremental stage for 15 epochs for TR3D, and follow SDCoT~\cite{zhao2022sdcot} for VoteNet.
We use the default optimizer and data augmentation settings provided by each backbone.
The memory bank maintains a fixed global budget of scenes, set to $10\%$ of the full training set. Each incremental stage is divided into $I{=}5$ sub-stages for learning-dynamics monitoring.
The learning dynamics recall $R^i_t(s,c)$ is computed at an IoU threshold of $0.5$.
The review emphasis strength is set to $\eta=3.0$ (Eq.~\ref{eq:replay_weight}).
The learnability-diversity trade-off parameter is $\lambda=0.5$ (Eq.~\ref{eq:bank_evolution}). 

\noindent\textbf{Evaluation protocol.}
We report mean average precision at IoU threshold 0.25 (mAP$\mathrm{@}$0.25) and Average Forgetting (AF) after every training stage.
After stage~$t$, mAP is computed over all seen classes $\mathcal{C}_{\leq t}$.
To quantify forgetting at stage~$t$, we let $\mathrm{AP}^{(t)}_{c}$ denote the AP$\mathrm{@}$0.25 of class~$c$ using the stage-$t$ model, and let $\mathrm{intro}(c)$ be the stage at which class~$c$ first appears.
AF measures the drop from each class's debut performance to its current performance (lower is better):
\begin{equation}
  \mathrm{AF}({\mathcal{C}_{<t}}) \;=\; \frac{1}{|\mathcal{C}_{<t}|} \sum_{c\,\in\,\mathcal{C}_{<t}}
  \Bigl(\mathrm{AP}^{(\mathrm{intro}(c))}_{c} - \mathrm{AP}^{(t)}_{c}\Bigr).
\label{af}
\end{equation}

% =========================================================
% Table 1a. Multi-stage class-incremental 3D detection (final stage mAP$\mathrm{@}$0.25)
% =========================================================
\begin{table}[!t]
\centering
\caption{Incremental 3D object detection performance (mAP$\mathrm{@}$0.25) over 3-/5-/10-incremental stages on SUN RGB-D and ScanNetV2 benchmarks.}
% \vspace{-0.3cm}
\label{tab:multistage_final}
\setlength{\tabcolsep}{4pt}
\renewcommand{\arraystretch}{1.05}
\begin{threeparttable}
\resizebox{\linewidth}{!}{%
\begin{tabular}{c c c c c | c c c}
\toprule[2pt]
\multirow{2}{*}{\textbf{Method}} & \multirow{2}{*}{\textbf{Backbone}}
& \multicolumn{3}{c|}{\textbf{SUN RGB-D ($40$ classes)}}
& \multicolumn{3}{c}{\textbf{ScanNetV2 ($35$ classes)}} \\
\cmidrule(lr){3-5}\cmidrule(lr){6-8}
& & \textbf{3-stage} & \textbf{5-stage} & \textbf{10-stage}
  & \textbf{3-stage} & \textbf{5-stage} & \textbf{10-stage} \\
\midrule
Fine-tuning
& \multirow{7}{*}{VoteNet~\cite{qi2019votenet}}
& 1.12 & 0.62 & 0.03
& 5.09 & 4.21 & 0.15 \\
CPDet3D~\cite{Zhu_2025_CVPR_CPDet3D}
&
& 9.70 & 7.11 & 0.39
& 16.33 & 9.05 & 2.54 \\
AIC3DOD~\cite{aic3dod}
&
& 10.73 & 7.32 & 0.85
& 19.21 & 15.51 & 1.60 \\
SDCoT~\cite{zhao2022sdcot}
&
& 10.94 & 7.66 & 0.98
& 18.39 & 14.38 & 2.93 \\
SDCoT++~\cite{zhao2024sdcot}
&
& 11.39 & 7.86 & 1.67
& 18.51 & 16.19 & 3.38 \\
SDCoT++ \& Random Memory
&
& 11.52 & 7.93 & 3.53
& 21.19 & 19.46 & 7.25 \\
\rowcolor{gray!10} \textbf{LDMR} (ours)
&
& \textbf{11.87} & \textbf{8.75} & \textbf{6.20}
& \textbf{24.82} & \textbf{23.37} & \textbf{9.76} \\
\midrule
\textit{Joint Training (Upper Bound)}
&
& 22.39 & 22.39 & 22.39
& 36.00 & 36.00 & 36.00 \\
\midrule
Fine-tuning
& \multirow{7}{*}{TR3D~\cite{tr3d2023}}
& 7.86 & 2.37 & 0.59
& 11.49 & 6.24 & 0.50 \\
CPDet3D~\cite{Zhu_2025_CVPR_CPDet3D}
&
& 8.71 & 7.55 & 3.32
& 18.76 & 7.26 & 0.66 \\
AIC3DOD~\cite{aic3dod}
&
& 22.43 & 15.37 & 9.11
& 32.69 & 20.11 & 8.66 \\
SDCoT~\cite{zhao2022sdcot}
&
& 22.10 & 13.77 & 8.72
& 34.65 & 20.86 & 6.93 \\
SDCoT++~\cite{zhao2024sdcot}
&
& 22.92 & 14.28 & 8.45
& 35.15 & 20.61 & 7.06 \\
SDCoT++ \& Random Memory
&
& 25.18 & 20.02 & 15.54
& 36.86 & 23.67 & 14.86 \\
\rowcolor{gray!10} \textbf{LDMR} (ours)
&
& \textbf{29.12} & \textbf{25.10} & \textbf{19.38}
& \textbf{39.00} & \textbf{28.38} & \textbf{17.82} \\
\midrule
\textit{Joint Training (Upper Bound)}
&
& 33.26 & 33.26 & 33.26
& 51.52 & 51.52 & 51.52 \\
\bottomrule[2pt]
\end{tabular}
}
\end{threeparttable}
% \vspace{-0.4cm}
\end{table}

\begin{figure}[!t]
  \centering
  \begin{minipage}[t]{0.48\linewidth}
    \centering
    \includegraphics[width=\linewidth]{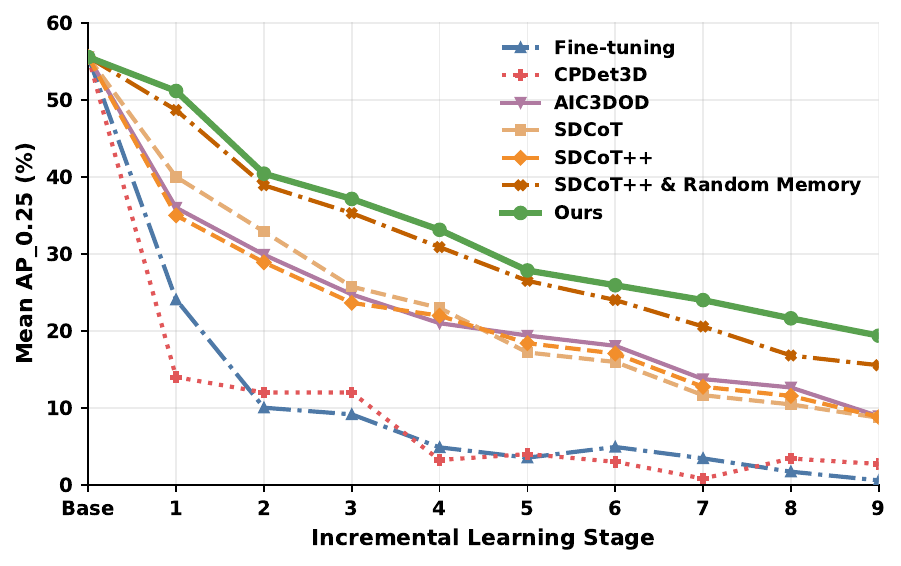}
    \label{fig:trajectory_ap25}
  \end{minipage}\hfill
  \begin{minipage}[t]{0.48\linewidth}
    \centering
    \includegraphics[width=\linewidth]{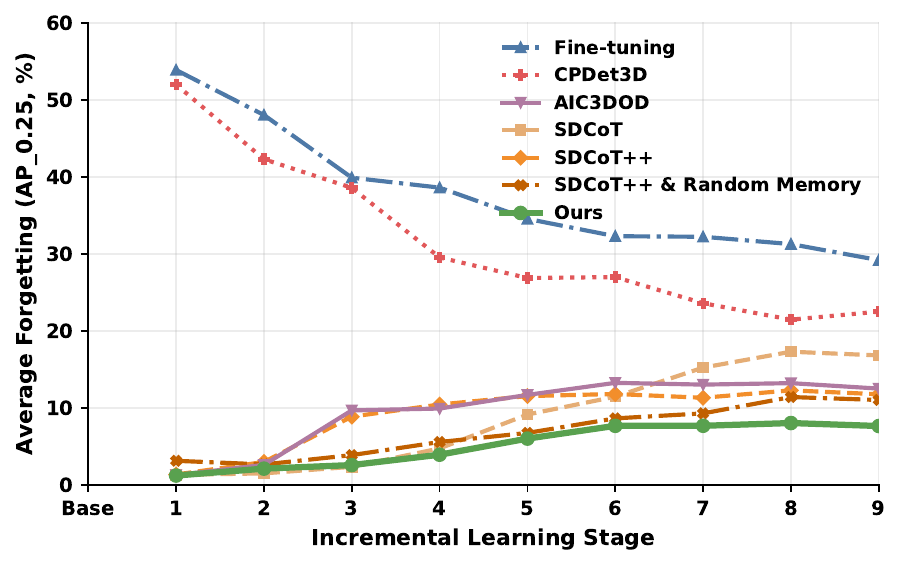}
    \label{fig:forgetting_ap25}
  \end{minipage}
  % \vspace{-0.6cm}
  \caption{Stage-wise experimental analysis on SUN RGB-D (with TR3D).
  \emph{Left:} Overall mAP on all seen classes after each stage.
  \emph{Right:} Average forgetting computed by Eq.~(\ref{af}) of old classes at each stage.
  Our method sustains higher seen-class performance and substantially reduces the forgetting issue across stages compared to all baselines.}
  \label{fig:stage_analysis}
  % \vspace{-0.1cm}
\end{figure}

\subsection{Baselines}
\label{sec:baselines}
We compare with the following baselines under the same protocol.
\textbf{Fine-tuning} is a fundamental baseline that trains only on current-stage ground-truth labels.
\textbf{CPDet3D}~\cite{Zhu_2025_CVPR_CPDet3D} is one of the state-of-the-art 3D detection methods which mines unlabeled objects via class prototypes and pseudo-label refinement; we adapt it to our incremental setting by treating old-class objects in new-stage scenes as the unlabeled targets.
\textbf{SDCoT}~\cite{zhao2022sdcot}, \textbf{SDCoT++}~\cite{zhao2024sdcot}, and \textbf{AIC3DOD}~\cite{aic3dod} are the representative incremental 3D object detection methods that rely on teacher-student training with pseudo labels and distillation to retain old-class knowledge.
No I3DOD methods are memory-bank based so far. Therefore we include \textbf{SDCoT++ \& Random Memory} which augments SDCoT++ with a randomly selected memory bank of the same budget as ours.

% =========================================================
% Table 1b. TR3D memory-only baselines (final stage mAP$\mathrm{@}$0.25)
% =========================================================

\subsection{Main Results}
\label{sec:main_results}

\noindent\textbf{Overall comparison.}
Tab.~\ref{tab:multistage_final} reports the final-stage's mAP$\mathrm{@}$0.25 for all protocols. Among pseudo-labeling based baselines, SDCoT++ is the strongest, but its performance degrades as the number of stages increases. For example, on SUN RGB-D (TR3D), it achieves 22.92 at 3 stages but only 8.45 at 10 stages. Augmenting SDCoT++ with a memory bank leads to a clear improvement, reaching 15.54 at 10 stages. 
% demonstrating the value of real ground-truth replay over pseudo labels alone.
Overall, our method LDMR consistently outperforms this strongest baseline across all settings. On SUN RGB-D (TR3D), we achieve 29.12, 25.10 and 19.38 at 3, 5 and 10 stages, improving over SDCoT++ \& Random Memory by 3.94, 5.08, and 3.84 mAP respectively. A similar pattern holds on ScanNetV2 (TR3D), where we reach 39.00, 28.38 and 17.82 compared to 36.86, 23.67, 14.86 by SDCoT++ \& Random Memory. These results suggest that our learning-dynamics-driven memory and review provide clear benefits beyond simply having a memory bank in pseudo-labeling methods. For reference, joint training on all classes reaches 33.26/51.52 mAP$\mathrm{@}$0.25 (SUN RGB-D/ScanNetV2, TR3D), serving as a non-incremental upper bound.

\noindent\textbf{Stage-wise trajectory and forgetting.}
Fig.~\ref{fig:stage_analysis} visualises the per-stage model performance on SUN RGB-D. The left panel shows that our method maintains the highest overall mAP after every stage, while baseline methods show increasingly larger drops as stages progress. The right panel shows that average forgetting of old classes remains consistently lower for our method across all stages, suggesting that our intra-stage review and cross-stage memory evolution effectively preserve previously learned knowledge.

\subsection{Ablation Studies}
\label{sec:ablation}
We further verify effectiveness of our method by detailed ablation studies.

\begin{table*}[!t]
\centering
\small
\renewcommand{\arraystretch}{1.0}
\setlength{\tabcolsep}{6pt}
% ===== 左侧表格 =====
\parbox{0.50\linewidth}{
\centering
\caption{{Ablation study on our key modules of human-like intra-stage review (HR) and scene-aware cross-stage memory evolution (ME) (SUN RGB-D, TR3D, 3-/5-/10-stages' mAP$\mathrm{@}$0.25)}.}
\label{tab:ablation_sun}
% \vspace{-0.2cm}
\resizebox{\linewidth}{!}{%
\begin{tabular}{l c c c}
\toprule
\textbf{Method} & \textbf{3-stage} & \textbf{5-stage} & \textbf{10-stage} \\
\midrule
\rowcolor{gray!10}
LDMR (full) & \textbf{29.12} & \textbf{25.10} & \textbf{19.38} \\
w/o HR & 26.67 & 23.47 & 17.51 \\
w/o ME & 27.44 & 23.17 & 18.70 \\
w/o HR \& ME & 25.18 & 20.02 & 15.54 \\
\bottomrule
\end{tabular}
}
}
\hfill
% ===== 右侧表格 =====
\parbox{0.44\linewidth}{
\centering
\caption{Comparison of alternative memory-bank schemes in SDCoT++ \& Random Memory (SUN RGB-D, TR3D, 10-stage's mAP$\mathrm{@}$0.25).}
\label{tab:tr3d_memory_only}
\setlength{\tabcolsep}{6pt}
% \vspace{-0.2cm}
\resizebox{\linewidth}{!}{%
\begin{tabular}{l c}
\toprule
\textbf{Memory-bank scheme} & \textbf{mAP$\mathrm{@}$0.25} \\
\midrule
Random            & 15.54 \\
Reservoir         & 16.12 \\
Max \#objects     & 15.31 \\
Lowest recall     & 14.10 \\
\rowcolor{gray!10} LDMR & \textbf{19.38} \\
\bottomrule
\end{tabular}
}
}
% \vspace{-0.2cm}
\end{table*}

\noindent\textbf{Intra-stage review \textit{vs}. cross-stage memory evolution.}
Tab.~\ref{tab:ablation_sun} demonstrates the individual contribution of key components in our LDMR framework.
On SUN RGB-D (TR3D), removing human-like intra-stage review (w/o HR) reduces 10-stage mAP from 19.38 to 17.51 and 
removing scene-aware memory evolution (w/o ME) reduces it to 18.70.
Removing both further drops the 10-stage mAP to 15.54, indicating that the two components are complementary, and their impact holds under different settings. At 3 stages the full model improves over the no-component baseline by 3.94 mAP, and at 5 stages the gain remains 5.08 mAP.

\noindent\textbf{Memory-bank selection schemes.}
To verify that the improvement does not merely come from using a memory bank, we equip the SDCoT++ \& Random Memory baseline with alternative memory-bank selection schemes (Tab.~\ref{tab:tr3d_memory_only}).
Reservoir sampling, max-object-count, and lowest-recall heuristics get 14.10--16.12 mAP, all below LDMR's 19.38, confirming that the gain stems from our design rather than from the bank itself.

\noindent\textbf{Scoring-function design choices.}
Tab.~\ref{tab:design_ablation} ablates scoring functions.
For the object-count weight in Eqs.~(\ref{eq:scene_drop}) and~(\ref{eq:scene_learnability}), the log form $\log(1{+}n_{s,c})$ (19.38) outperforms the linear $n_{s,c}$ (18.55) and sublinear $\sqrt{n_{s,c}}$ (19.01) forms, as strong regulation best prevents object-dense scenes from dominating the score. 
Moreover, each term in the learnability score (Eq.~\ref{eq:scene_learnability}) is necessary: removing the under-learning weight $\omega_t(c)$, the count weight $\log(1{+}n_{s,c})$, or the learning gain $G_t(s,c)$ reduces performance to 19.13, 17.86, and 18.88, respectively. 
Besides, simplifying the under-learning weight (Eq.~\ref{eq:class_need}) to a low-recall only form drops performance.

\noindent\textbf{Adaptive \textit{vs}.\ fixed reviewing.}
Fig.~\ref{fig:param_sensitivity_a} compares adaptive-weight reviewing against fixed-weight reviewing on the 10-stage protocol.
Our approach outperforms the fixed-weight variant, showing that the learning-dynamics-driven adaptation effectively allocates review effort as forgetting patterns vary across stages.

\noindent\textbf{Reviewing granularity.}
Fig.~\ref{fig:param_sensitivity_b} examines the effect of the reviewing granularity, i.e., the number of sub-stages $I$ into which each incremental stage is divided.
Each stage is split by $I{-}1$ review checkpoints into $I$ sub-stages (e.g., a single checkpoint for two sub-stages).
Across all incremental protocols, performance clearly improves as the number of review checkpoints increases from 0 ($I$ increases from 1), indicating that finer-grained monitoring of learning dynamics enables more targeted review.
Meanwhile, performance plateaus for too many sub-stages, suggesting that the reviewing segments should not be too short.

\begin{table}[!t]
\centering
\caption{Ablation on the scoring-function design choices (SUN RGB-D, TR3D, 10-stage mAP$\mathrm{@}$0.25). Each variant alters one component of LDMR.}
\label{tab:design_ablation}
% \vspace{-0.2cm}
\setlength{\tabcolsep}{6pt}
\renewcommand{\arraystretch}{1.05}
\resizebox{\linewidth}{!}{%
\begin{tabular}{l l c}
\toprule
\textbf{Component} & \textbf{Design choice} & \textbf{mAP$\mathrm{@}$0.25} \\
\midrule
\multirow{3}{*}{Count weight (Eqs.~\ref{eq:scene_drop},\ref{eq:scene_learnability})}
 & linear: $n_{s,c}$            & 18.55 \\
 & sublinear: $\sqrt{n_{s,c}}$  & 19.01 \\
\rowcolor{gray!10}
 & log: $\log(1+n_{s,c})$ (final design) & \textbf{19.38} \\
\midrule
\multirow{3}{*}{Learnability (Eq.~\ref{eq:scene_learnability})}
 & w/o $\omega_t(c)$ (class under-learning weight)   & 19.13 \\
 & w/o $\log(1+n_{s,c})$ (object-count weight)       & 17.86 \\
 & w/o $G_t(s,c)$ (scene-level learning gain)        & 18.88 \\
\midrule
Under-learning weight (Eq.~\ref{eq:class_need}) & low-recall only & 18.45 \\
\bottomrule
\end{tabular}
}
\end{table}

\begin{figure}[!t]
    \centering
    \begin{subfigure}[t]{0.22\linewidth}
        \centering
        \includegraphics[width=\linewidth]{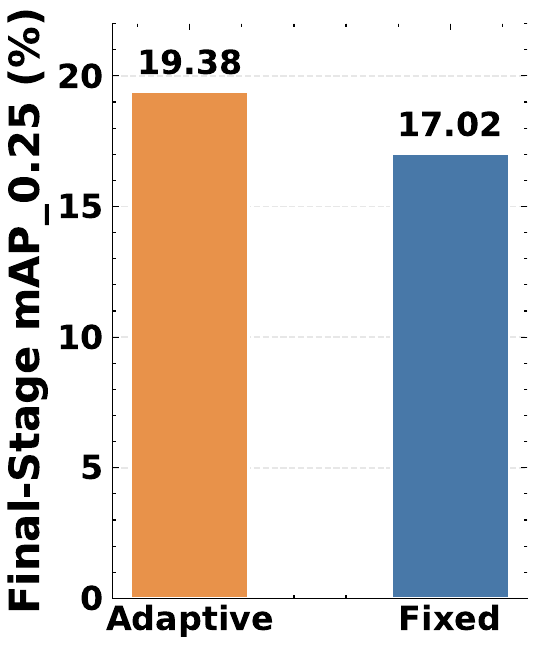}
        \caption{Reviewing weight.}
        \label{fig:param_sensitivity_a}
    \end{subfigure}\hfill
    \begin{subfigure}[t]{0.37\linewidth}
        \centering
        \includegraphics[width=\linewidth]{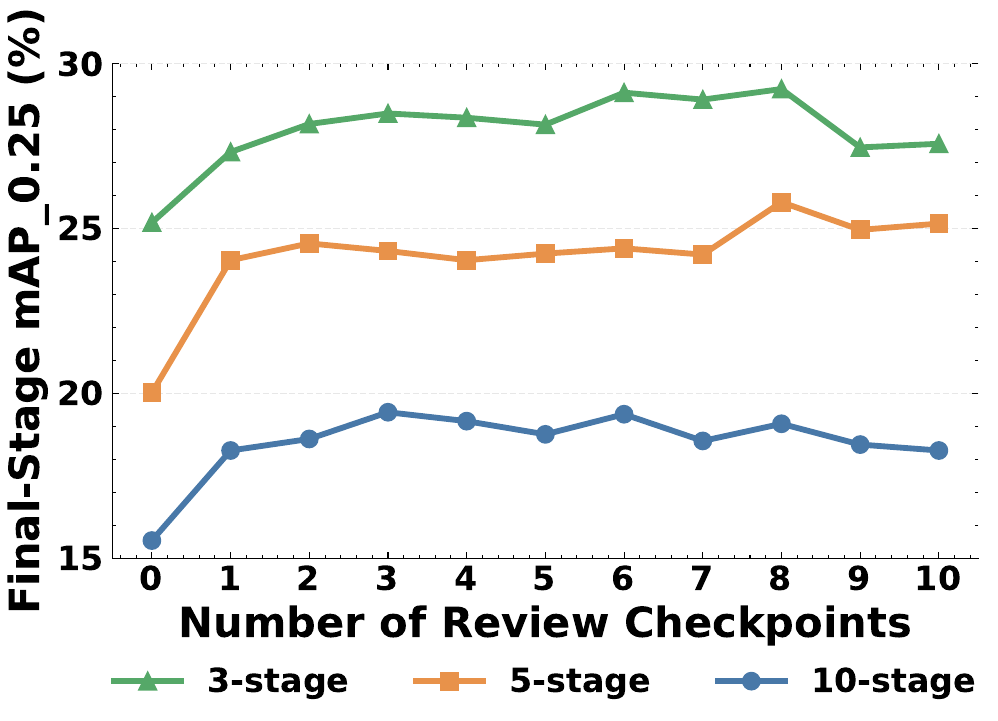}
        \caption{Number of review checkpoints.}
        \label{fig:param_sensitivity_b}
    \end{subfigure}\hfill
    \begin{subfigure}[t]{0.37\linewidth}
        \centering
        \includegraphics[width=\linewidth]{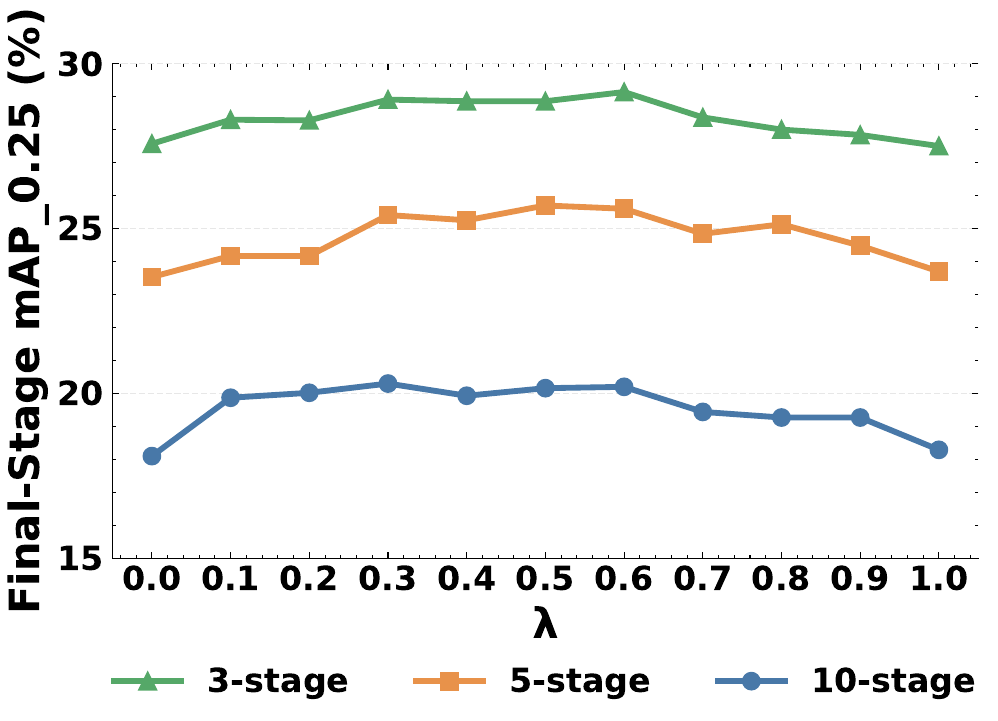}
        \caption{Trade-off $\lambda$.}
        \label{fig:param_sensitivity_c}
    \end{subfigure}
    % \vspace{-0.1cm}
   \caption{
    (a) Ablation study on adaptive and fixed sampling weights in the human-like intra-stage review on SUN RGB-D (with TR3D).
    (b) Ablation study on the reviewing granularity.
    (c) Effect of the trade-off parameter $\lambda$ between learnability and diversity.}
    \label{fig:param_sensitivity}
    % \vspace{-0.3cm}
\end{figure}

\noindent\textbf{Trade-off between learnability and diversity.}
Fig.~\ref{fig:param_sensitivity_c} studies the effect of the trade-off parameter $\lambda$ in the memory evolution objective (Eq.~\ref{eq:bank_evolution}).
Setting $\lambda{=}0$ (learnability only, no diversity) or $\lambda{=}1$ (diversity only, no learnability) generally results in lower performance, while the middle value of $\lambda$ improves results across all protocols, with the best performance reached around $\lambda{=}0.3 \sim 0.6$.
Beyond this range, performance gradually declines, indicating that insufficient learnability or diversity hurts performance.
The result suggests that both terms are necessary and that a moderately balanced setting works best.

\section{Conclusion}
\label{sec:con}

In this work, we studied long-incremental 3D object detection and showed that existing pipelines can enter a self-reinforcing cycle, where stage-wise supervision shift degrades old-class performance and progressively corrupts pseudo labels. To break this cycle, we proposed Learning-Dynamics-driven Memory and Review (LDMR), which converts learning dynamics into training decisions through human-like intra-stage review and scene-aware cross-stage memory evolution. Across SUN RGB-D and ScanNetV2 under multiple incremental protocols, LDMR consistently improves the performance by mitigating model forgetting, with more pronounced gains in longer incremental stages, highlighting the importance of our learning-dynamics-driven framework for long-horizon 3D continual learning.

\section{Acknowledgements}
This research is supported by the Agency for Science, Technology and Research (A*STAR) under its MTC Programmatic Funds (Grant No. M23L7b0021), and the Ministry of Education, Singapore, under its MOE Academic Research Fund Tier 2 (MOE-T2EP20124-0013).

%----14 pages----

% \section*{Acknowledgements}
% Please insert your acknowledgments here.

% \newpage

\bibliographystyle{splncs04}
\bibliography{main}
\end{document}